\documentclass[10pt,logo,copyright]{nvidiatechreport}
\usepackage[authoryear,round]{natbib}

\usepackage[utf8]{inputenc} 
\usepackage[T1]{fontenc}    

\usepackage{parskip}        
\usepackage{url}            
\usepackage{booktabs}       
\usepackage{amsfonts}       
\usepackage{nicefrac}       
\usepackage{microtype}      
\usepackage{xcolor}         
\usepackage[dvipsnames]{xcolor} 
\usepackage{graphicx}
\usepackage{animate}        
\usepackage{subcaption}
\usepackage{tabularx}
\usepackage{makecell}
\usepackage{adjustbox}
\usepackage{setspace}
\newcolumntype{M}[1]{>{\centering\arraybackslash}m{#1}}
\usepackage{float}
\usepackage{tikz}
\usetikzlibrary{positioning,shapes,arrows}
\usepackage{amsmath,amsfonts,bm, bbm,leftindex}
\usepackage{multirow}
\usepackage{comment}
\usepackage{gensymb}
\usepackage{lipsum}
\usetikzlibrary{arrows.meta, positioning, fit}
\usepackage[para]{threeparttable}
\usepackage{tikz}
\usetikzlibrary{tikzmark}

\usepackage[most]{tcolorbox}
\usepackage{fancyvrb}
\usepackage{fvextra}
\usepackage{dashrule}

\usepackage{pifont}

\usepackage[nameinlink]{cleveref}
\crefname{equation}{Eq.}{Eqs.}
\crefname{figure}{Fig.}{Figs.}
\crefname{section}{Sec.}{Sec.}
\crefname{appendix}{App.}{App.}
\crefname{table}{Tab.}{Tabs.}
\crefname{algorithm}{Algo}{Algo}
\crefname{thm}{Thm}{Thm}
\Crefname{thm}{Thm}{Thm}
\crefname{prop}{Prop}{Prop}

\definecolor{darkred}{rgb}{0.7, 0.0, 0.0}

\def\ours{ThinkAct}

\newcommand{\crefnames}[3]{%
  \@for\next:=#1\do{%
    \expandafter\crefname\expandafter{\next}{#2}{#3}%
  }%
}

\title{ThinkAct: Vision-Language-Action Reasoning via Reinforced Visual Latent Planning}

\author{
    {Chi-Pin Huang}$^{1,2}$\quad {Yueh-Hua Wu}$^1$\quad {Min-Hung Chen}$^1$ \quad {Yu-Chiang Frank Wang}$^{1,2}$\quad {Fu-En Yang}$^{1}$\\
    \normalsize \textsuperscript{1} NVIDIA \quad
    \normalsize \textsuperscript{2} National Taiwan University
}

\begin{document}

\maketitle

\begin{abstract}
Vision-language-action (VLA) reasoning tasks require agents to interpret multimodal instructions, perform long-horizon planning, and act adaptively in dynamic environments. Existing approaches typically train VLA models in an end-to-end fashion, directly mapping inputs to actions without explicit reasoning, which hinders their ability to plan over multiple steps or adapt to complex task variations. In this paper, we propose ThinkAct, a dual-system framework that bridges high-level reasoning with low-level action execution via reinforced visual latent planning. ThinkAct trains a multimodal LLM to generate embodied reasoning plans guided by reinforcing action-aligned visual rewards based on goal completion and trajectory consistency. These reasoning plans are compressed into a visual plan latent that conditions a downstream action model for robust action execution on target environments. Extensive experiments on embodied reasoning and robot manipulation benchmarks demonstrate that ThinkAct enables few-shot adaptation, long-horizon planning, and self-correction behaviors in complex embodied AI tasks. Project Page: \textnormal{\href{https://jasper0314-huang.github.io/thinkact-vla/}{\textcolor{LimeGreen}{https://jasper0314-huang.github.io/thinkact-vla/}}}

\end{abstract}
\abscontent
\section{Introduction}
\label{sec:intro}

Recent advances in multimodal large language models (MLLMs)~\cite{team2024gemini,liu2023visual,bai2025qwen2,shi2024eagle,lin2024vila,achiam2023gpt,li2024llava,chen2024expanding,liu2024nvila,zhu2025internvl3,li2025eagle,chen2025eagle} have led to impressive progress on various tasks requiring the understanding of multimodal inputs, such as visual question answering and image/video captioning. However, while multimodal content can now be effectively perceived and interpreted, conducting multi-step planning for long-horizon user goals and then interacting with dynamic environments remains challenging for frontier MLLMs. Therefore, enabling the vision-language foundation models with action awareness and embodied reasoning capabilities unleashes a wide range of physical AI applications (e.g., robotics and AR assistance), and draws significant attention from both academics and industry.    

To bridge action with vision-language modalities, several works~\cite{brohan2023rt,kim24openvla,zheng2024tracevla,bjorck2025gr00t,team2024octo} learn vision-language-action (VLA) models by initializing from pre-trained MLLMs and training on large-scale robotic demonstrations (e.g., Open X-Embodiment Dataset~\cite{o2024open}). For example, OpenVLA~\cite{kim24openvla} builds upon MLLMs with post-training on large-scale robot demonstrations, while TraceVLA~\cite{zheng2024tracevla} further applies visual traces prompting to enhance spatial context understanding. Despite promising on short-horizon skills, the crucial capabilities to reason in diverse visual scenes and enable long-horizon planning remain limited due to the \textit{end-to-end} fashion from visual and textual inputs to low-level actions.

\begin{figure}[t!]
    \centering
    \includegraphics[width=1.0\linewidth]{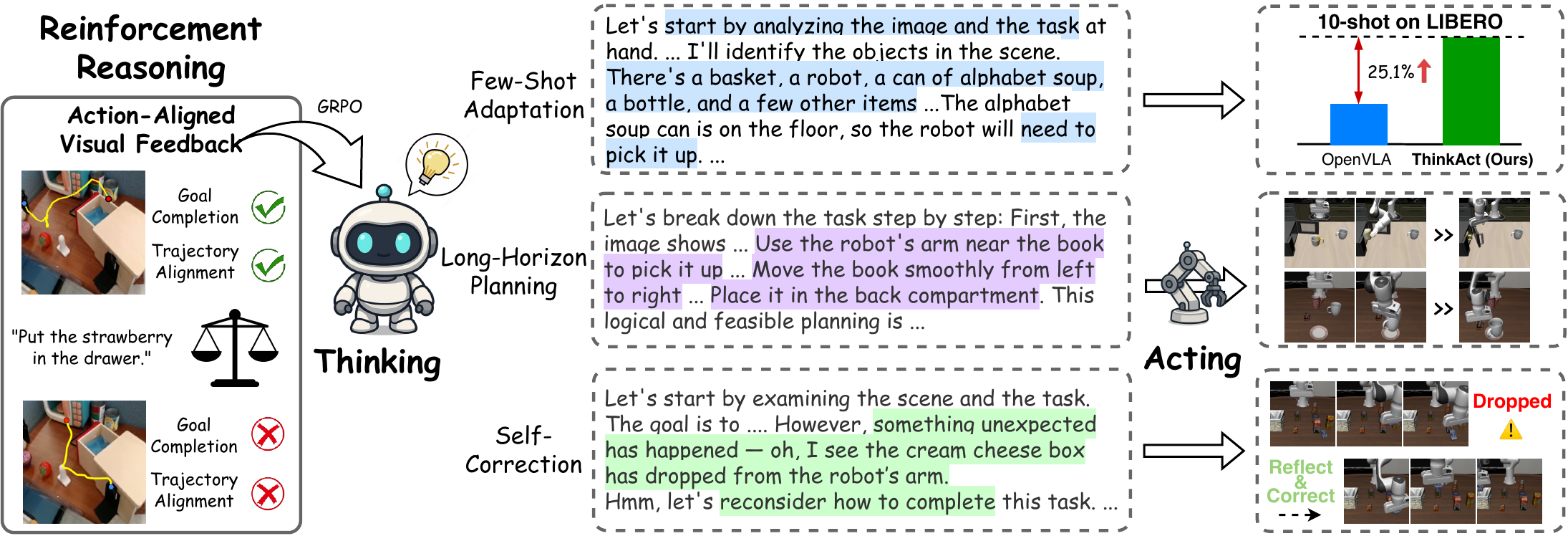}
    \caption{We introduce \ours{}, a reasoning VLA framework capable of thinking before acting. Through reasoning reinforced by our \emph{action-aligned visual feedback}, \ours{} enables capabilities of few-shot adaptation, long-horizon planning, and self-correction in embodied tasks.}
    \label{fig:teaser}
\end{figure}

To equip VLAs with the ability to solve complex embodied tasks, recent works~\cite{zawalski2024robotic,clark2025action,zhao2025cot,shi2025hi} have explored incorporating explicit chain-of-thought (CoT) prompting~\cite{wei2022chain} as an intermediate step-by-step guidance. For instance, ECoT~\cite{zawalski2024robotic} and RAD~\cite{clark2025action} introduce data curation pipelines to generate intermediate steps and decomposed plans by prompting off-the-shelf MLLMs. Once the annotated CoT traces are obtained, VLAs are trained to predict intermediate steps via fully \textit{supervised fine-tuning (SFT)}. However, due to the high cost of producing high-quality reasoning traces, the resulting models are prone to overfitting to specific visual scenes or reasoning patterns.

Recently, reinforcement learning (RL)~\cite{shao2024deepseekmath,guo2025deepseek} has demonstrated significant potential to incentivize reasoning behaviors in LLMs by exploring the thinking trace that maximizes reward signals instead of solely relying on fully supervised CoT annotations. Inspired by this paradigm, several vision-language models~\cite{feng2025video,nvidia2025cosmosreason1physicalcommonsense,tan2025reason} have applied RL-based reasoning to multimodal tasks. For example, Video-R1~\cite{feng2025video} adopts R1-style RL optimization to induce the CoT traces by verifiable answer accuracy with format correctness. While this manner enables long-form reasoning without step-level supervision, the reliance on QA-style reward signals limits their ability to support long-horizon planning and makes it difficult to connect reasoning with real-world action execution.

In this paper, we propose \textit{\ours{}}, which aims to enable MLLMs with the capability to reason before acting in physical environments. To address vision-language-action reasoning tasks, \ours{} adopts a dual-system architecture that connects structured reasoning with executable actions. Specifically, we incentivize MLLMs to perform long-horizon planning by advancing reinforcement learning with an action-aligned reward, derived from visual goal completion and trajectory distribution matching. Our \ours{} leverages human and robot videos to elicit embodied reasoning that is grounded in visual observations. To bridge reasoning and execution, we compress intermediate reasoning steps into a compact latent trajectory that captures high-level intent and allows efficient adaptation of the downstream action network to new environments. By reinforcing structured reasoning and grounding it in real-world actions, \ours{} tackles long-horizon manipulation tasks while unleashing few-shot action adaptation and self-correction behavior in physical AI scenarios, as shown in Fig.~\ref{fig:teaser}.

Our main contributions are summarized as follows:
\begin{itemize}
    \item We propose \textit{\ours{}}, a dual-system framework that mutually enhances action execution and visual-grounded embodied reasoning connected by visual latent planning.

    \item We leverage the visual feedback of goal completion and trajectory alignment as action-aligned rewards to allow long-horizon reasoning grounded in the embodied scene.
    
    \item We advance visual latent planning to steer downstream action execution by providing reasoning-enhanced trajectory guidance across diverse environments.
    
    \item We demonstrate that our learned reasoning VLA enables capabilities of few-shot adaptation, long-horizon planning, and self-correction across diverse embodied manipulation tasks.
\end{itemize}
\section{Related Works}
\label{sec:related_works}

\subsection{Vision-Language-Action Models}

Recent efforts~\cite{li2024llara,yuan2024robopoint,duan2024aha,niu2024llarva} have adapted vision-language models (VLMs) for action-centric tasks by post-training on curated instruction-following data. For example, RoboPoint~\cite{yuan2024robopoint} and LLARVA~\cite{niu2024llarva} leverage point and visual trajectory into textual prompts to augment LLMs with spatial-action understanding ability. AHA~\cite{duan2024aha} enhances failure detection ability in robotic manipulation by formulating it as a free-form question-answering task, training on synthetic failure data generated by perturbing successful trajectories. Although effective in specific domains, these approaches depend on sophisticatedly curated data and struggle to generalize beyond their training distributions. To improve scalability, recent vision-language-action (VLA) models~\cite{kim24openvla,zheng2024tracevla,szot2024multimodal,bjorck2025gr00t,li2025hamster,yang2025magma,brohan2022rt} adopt large-scale robot datasets (e.g., Open X-Embodiment Dataset~\cite{o2024open} or DROID~\cite{khazatsky2024droid}) to train models directly on diverse demonstrations. OpenVLA~\cite{kim24openvla} learns from pre-trained VLMs with robot trajectories for generalist action execution, while TraceVLA~\cite{zheng2024tracevla} and HAMSTER~\cite{li2025hamster} enhance spatial-action awareness by incorporating visual traces. However, these models predict actions directly from vision and language inputs, often bypassing structured planning or intermediate reasoning. As a result, their capability to handle complex instructions, long-horizon goals, or out-of-distribution scenarios remains limited.

\subsection{Reasoning in Vision-Language-(Action) Models}

Chain-of-thought (CoT) prompting~\cite{wei2022chain,wang2024chain,yeo2025demystifying} has significantly improved the multi-step reasoning ability of LLMs across math, coding, and question-answering tasks. Motivated by these advances, recent works extend reasoning capabilities to vision-language-action (VLA) models for embodied tasks.
ECoT~\cite{zawalski2024robotic} synthesizes intermediate subgoals via prompting and applies supervised fine-tuning to teach VLAs to reason before acting. RAD~\cite{clark2025action} leverages action-free human videos to curate reasoning traces by prompting off-the-shelf LLMs and learn to map reasoning to real actions using robot data. On the other hand, CoT-VLA~\cite{zhao2025cot} replaces linguistic CoT with visual subgoal frames generated ahead of action prediction. However, they depend on either curated CoT supervision or task-specific video generation, limiting their scalability. Inspired by the recent success of RL-optimized reasoning models~\cite{shao2024deepseekmath,guo2025deepseek}, several approaches~\cite{feng2025video,nvidia2025cosmosreason1physicalcommonsense,tan2025reason,liu2025visual} adopt GRPO~\cite{shao2024deepseekmath} optimization to guide CoT generation in vision-language tasks using verifiable rewards. However, their QA-formatted rewards cannot fully support long-horizon planning or establish grounding between reasoning and action execution. To unify structured CoT reasoning with embodied decision-making, we introduce \ours{}, which leverages action-aligned reinforcement learning and visual latent planning to connect embodied reasoning with real-world action in VLA tasks.
\section{Method}
\label{sec:method}

\subsection{Problem Formulation}

We first define the setting and notations for vision-language-action (VLA) reasoning tasks. At each timestep $t$, the model receives a visual observation $o_t$ and a textual instruction $l$, with the goal of predicting an action $a_t$, which can be a textual command or a 7-DOF control vector $\left[ \Delta_x, \Delta_\theta, \Delta_\text{Grip} \right]$ depending on the embodiment.
To tackle this problem, we propose \textit{\ours{}}, a unified framework that aims to leverage an MLLM $\mathcal{F}_\theta$ to reason the high-level plans while connecting with an action model $\pi_\phi$ to infer executable actions. The MLLM $\mathcal{F}_\theta$ produces a visual plan latent $c_t$ based on $(o_t, l)$, capturing the high-level intent and planning context (Sec.~\ref{ssec:ER}). This reasoned plan $c_t$ then guides the downstream action module $\pi_\phi$ to sequentially predict $N$ executable actions $[a_t]_{t}^{t+N}$ tailored to the target environment (Sec.~\ref{ssec:action}). By connecting abstract planning with low-level control, our \ours{} enables long-horizon reasoning and improves action adaptation in dynamic embodied tasks.

\subsection{Reinforced Visual Latent Planning for Embodied Reasoning}
\label{ssec:ER}

\begin{figure}[t!]
    \centering
    \includegraphics[width=1.0\linewidth]{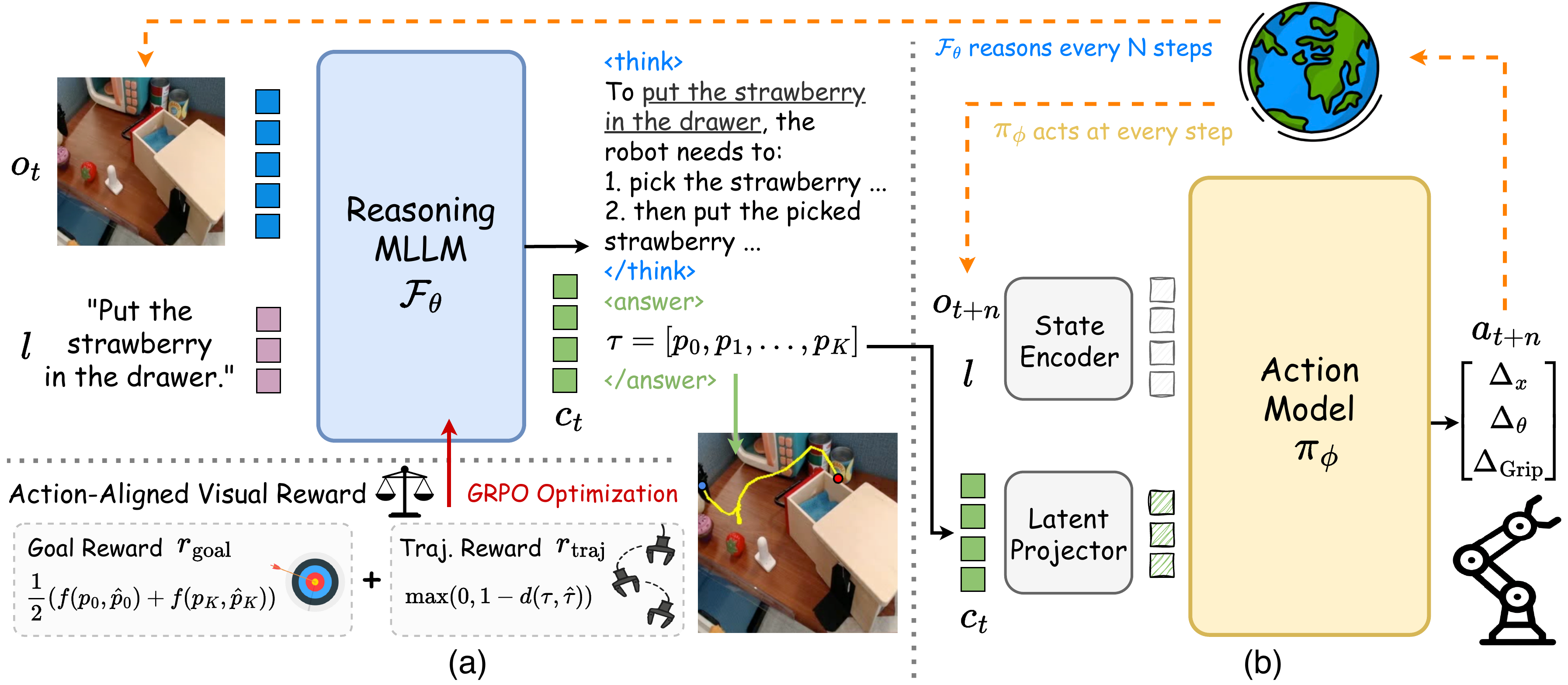}
    \caption{\textbf{Overview of our \ours{}.} (a) Given observation $o_t$ and instruction $l$, \ours{} advances \emph{action-aligned} rewards derived from visual trajectory $\tau$ to incentivize embodied reasoning capability of Reasoning MLLM $\mathcal{F}_\theta$. (b) Conditioned on the visual plan latent $c_t$, the DiT-based Action Model $\pi_\phi$ learns to predict executable action while keeping $\mathcal{F}_\theta$ frozen. Note that, during inference, $\pi_\phi$ and $\mathcal{F}_\theta$ could operate asynchronously to enable slow thinking and fast control for VLA reasoning tasks.}
    \label{fig:overview}
\end{figure}

To enable embodied reasoning that generalizes across diverse environments, we aim to incentivize the reasoning capability of multimodal LLMs via reinforcement learning~\cite{shao2024deepseekmath,guo2025deepseek}. A straightforward way is to have the MLLM reason before generating low-level actions, while using the resulting task success rate in target environments (e.g., LIBERO~\cite{liu2023libero}) as the reward signal. However, this approach is restricted to specific simulators without proper guidance from visual scenes.

\paragraph{Reward Shaping from Action-Aligned Visual Feedback}
To tackle this challenge, we design a novel action-aligned visual feedback that captures long-horizon goals and encourages visual grounding during planning. Specifically, inspired by recent works~\cite{yang2025magma,zheng2024tracevla}, we are capable of representing high-level plans as spatial-temporal trajectories that capture the gripper end-effector over the visual scene, which serve as a visual-action guidance to steer the embodied reasoning.

As depicted in Fig.~\ref{fig:overview}(a), given an observation $o_t$ at timestep $t$ and a task instruction $l$, the MLLM $\mathcal{F}_\theta$ autoregressively generates a sequence of latent embeddings for reasoning $v_t \in \mathbb{R}^{|v_t| \times d}$ and visual plan $c_t \in \mathbb{R}^{|c_t| \times d}$, where the former is decoded to reasoning steps while the latter would be inferred into a text string of 2D points $\tau = \left[p_k\right]_{k=1}^K$, with $p_k \in [0, 1]^2$, and $p_1$ and $p_K$ denoting the \emph{start} and \emph{end} positions of the gripper.
As a result, to encourage the model to anticipate visual goal completetion, we introduce the \emph{goal reward} for comparing  predicted start and end positions with corresponding points from trajectory obtained by off-the-shelf detector~\cite{niu2024llarva} $\hat{\tau} = \left[\hat{p}_k\right]_{k=1}^K$ as follows,
\begin{equation}
    r_{\text{goal}} = \frac{1}{2} \left( f\left( p_1, \hat{p}_1 \right) + f\left( p_K, \hat{p}_K \right) \right), \quad \text{where } f(p, p^\prime) = \max\left( 0, 1-\| p - p^\prime \|_2^2 \right).
\end{equation}

To further enforce the MLLM predicted trajectory to properly correspond to physically plausible gripper motion, the \emph{trajectory reward} is proposed to regularize the predicted $\tau$ to match the distribution of demonstrated trajectory $\hat{\tau}$. Thus, the trajectory reward $r_{\text{traj}}$ can be computed as follows,
\begin{equation}
    r_{\text{traj}} = \max\left( 0, 1 - d(\tau, \hat{\tau}) \right).
\end{equation}
Here, $d(\tau, \hat{\tau})$ denotes a metric measuring the distance between two trajectories, i.e., dynamic time warping (DTW) distance~\cite{senin2008dynamic} in this work.

The overall reward is thus defined as the combination of our proposed action-aligned visual feedback and the format correctness score $r_\text{format}$ following existing reasoning works~\cite {guo2025deepseek}:
\begin{equation}\label{eq:reward_total}
    r = 0.9 r_\text{visual} + 0.1 r_\text{format}, \text{where }r_\text{visual} = \omega_\text{goal}r_\text{goal} + \omega_\text{traj}r_\text{traj}.
\end{equation}
Here, $\omega_\text{goal} = \omega_\text{traj} = 0.5$ are the weighting coefficients for the goal and trajectory rewards.

\paragraph{Reinforced Fine-Tuning for Eliciting Visual Latent Planning}

To incentivize the embodied reasoning from the MLLM $\mathcal{F}_\theta$, we perform reinforced fin-tuning using Group Relative Policy Optimization (GRPO)~\cite{shao2024deepseekmath}.
Specifically, given an input $(o_t, l)$, GRPO first samples a group of $M$ distinct responses $\{ z_1, z_2, \ldots, z_M \}$ from the original MLLM $\mathcal{F}_{\theta_\text{old}}$. Each response is evaluated using the reward function defined in Eq.~\ref{eq:reward_total} and resulting in a set of reward signals $\{ r_1, r_2, ..., r_M \}$. Thus, we optimize $\mathcal{F}_{\theta}$ by maximizing the following objective:
\begin{equation}
    \mathcal{J}_\text{GRPO}(\theta) = \frac{1}{M} \sum_{i=1}^{M}(\frac{\mathcal{F}_{\theta}(z_i|o_t,l)}{\mathcal{F}_{\theta_\text{old}}(z_i|o_t,l)}A_i - \beta D_{KL}(\mathcal{F}_{\theta}(z_i|o_t,l)\parallel \mathcal{F}_{\theta_\text{old}}(z_i|o_t,l))),
\end{equation}
\begin{equation*}
     \\ \text{where} \quad A_i=\frac{r_i - \text{mean}(\{ r_1, \ldots, r_M \})}{\text{std}(\{ r_1, \ldots, r_M \})}.
\end{equation*}
Here, $A_i$ quantifies the relative quality of $i$-th response compared to other candidates in the sampled group. $D_{KL}(\cdot\parallel\cdot)$ is the KL divergence introduced with a weighting factor $\beta$ to regularize the model, preventing excessive deviation from the original model $\mathcal{F}_{\theta_\text{old}}$.

To further obtain general embodied knowledge, our \ours{} is flexible to encapsulate the publicly available question-answering data to enhance capabilities such as robotic VQA~\cite{sermanet2024robovqa} or failure detection~\cite{liu2023reflect} by formatting them into the QA-style accuracy reward. Once the reinforced fine-tuning is complete, we are able to produce long CoT steps, while abstracting the textual reasoning into a compact visual plan latent $c_t$, capturing long-horizon spatial-temporal planning intent.

\subsection{Reasoning-Enhanced Action Adaptation}
\label{ssec:action}

With the high-level embodied intent reasoned by the MLLM, our goal is to connect the inferred visual latent planning $c_t$ with the action model $\pi_\phi$ of the target environment in a think-before-acting manner, grounding embodied reasoning into the physical world with executable actions.
Specifically, we build upon a Transformer-based action model $\pi_\phi$ (e.g., Diffusion Policy~\cite{chi2023diffusion}), which predicts actions based on the current state composed of visual observations and language instructions. While $\pi_\phi$ can operate in the target environment using perception alone, we enhance its capability by conditioning it on the latent plan $c_t$, which encodes high-level embodied intent and planning context.

As depicted in Fig.~\ref{fig:overview}(b), we incorporate $c_t$ using a latent projector to connect it to the input space of the action model, enabling the reasoning guidance to be effectively leveraged, which enhances its low-level action execution in the target environment. Thus, we solely update the state encoder, latent projector, and action model by imitation learning with annotated action demonstrations: 
\begin{equation} \mathcal{L}_{\text{IL}}(\phi) = \mathbb{E}_{(o_i, l, a_i)} \left[ \ell\left(\pi_\phi(c_{t}, o_i, l), a_i \right) \right]. \end{equation} 

We note that, reasoning and action execution could be operated in an \textit{asynchronous} manner, which means each latent plan $c_t$ corresponds to $N$ interactions with the environment (i.e., $i\in[t, t+N]$). This asynchronous design highlights a key advantage of our dual-system architecture, allowing the reasoning MLLM to perform slow thinking while the action model executes fast control.

\subsection{Learning Strategy and Inference}

Following~\cite{feng2025video}, we adopt a multi-stage training strategy for our \ours{}. Before RL, we initialize the two modules independently. The MLLM $\mathcal{F}_\theta$ is cold-started using supervised data (Sec.~\ref{ssec:exp_setup}) to learn to interpret visual trajectories and produce reasoning and answers in the correct output format. On the other hand, the action model $\pi_\phi$ is pre-trained on the Open X-Embodiment (OXE) dataset~\cite{o2024open}, providing a strong foundation for low-level action execution.
After SFT cold-start, our MLLM $\mathcal{F}_\theta$ is tuned with action-aligned rewards guiding the generation of effective latent plans. During reasoning-enhanced action adaptation, we freeze $\mathcal{F}_\theta$ while updating the action model $\pi_\phi$ with state encoder and latent projector on the target environment by conditioning on the latent visual plan $c_t$.

At inference time, given a visual observation $o_t$ and instruction $l$, \ours{} produces a visual plan latent $c_t = \mathcal{F}_\theta(o_t, l)$, which conditions the action module $\pi_\phi$ to predict a sequence of executable actions tailored to the current environment.
\section{Experiment}
\label{sec:experiments}

\subsection{Experimental Setup}
\label{ssec:exp_setup}

\paragraph{Implementation Details}

We initialize $\mathcal{F}_\theta$ with Qwen2.5-VL 7B~\cite{bai2025qwen2}. The cold-start stage runs for 20K iterations with batch size 32 and learning rate $1\text{e}{-5}$ using DeepSpeed ZeRO-3. We then apply GRPO~\cite{shao2024deepseekmath} for 6K iterations, using batch size 64, learning rate $1\text{e}{-6}$, and rollout size 5. The action model $\pi_\phi$ is a DiT-based policy~\cite{chi2023diffusion} with 432M parameters, pre-trained using the OXE dataset~\cite{o2024open}, where the state encoder is composed of a DINOv2 image encoder~\cite{oquab2023dinov2} and a CLIP text encoder~\cite{radford2021learning} that jointly encode the current state inputs into $1024$-dim embeddings. For reasoning-enhanced action adaptation, we connect the visual plan $c_t$ via a Q-Former~\cite{li2023blip} as the latent projector with 32 queries and fine-tune on 100K OXE samples for 120K iterations using batch size 256 and learning rate $2\text{e}{-5}$. LIBERO~\cite{liu2023libero} tasks are further fine-tuned for 75K iterations with batch size 128. All experiments are conducted on 16 NVIDIA A100 GPUs with 80 GB memory.

\paragraph{Training Datasets and Evaluation Benchmarks}

For SFT cold-start, we fine-tune the MLLM using trajectories from the subset of OXE, and QA tasks from RoboVQA~\cite{sermanet2024robovqa}, EgoPlan-IT~\cite{chen2023egoplan}, and Video-R1-CoT~\cite{feng2025video}. During RL training, we incorporate trajectories from the OXE subset and human videos from Something-Something v2~\cite{goyal2017something}. To enhance general reasoning capability, we include embodied QA datasets such as EgoPlan-IT/Val~\cite{chen2023egoplan}, RoboVQA~\cite{sermanet2024robovqa}, and the Reflect dataset~\cite{liu2023reflect}, as well as a general video instruction dataset, i.e., LLaVA-Video-178K~\cite{zhang2024video}.

We evaluate \ours{} on two robot manipulation and three embodied reasoning benchmarks. For manipulation tasks, SimplerEnv~\cite{li24simpler} containing diverse scenes and LIBERO~\cite{liu2023libero} with long-horizon tasks are evaluated using task success rate. For reasoning benchmarks, EgoPlan-Bench2~\cite{qiu2024egoplan} uses accuracy on multiple-choice questions, while RoboVQA~\cite{sermanet2024robovqa} and OpenEQA~\cite{majumdar2023openeqa} are free-form QA tasks evaluated using BLEU score~\cite{papineni2002bleu} and LLM-based scoring, respectively, following their original protocols. Further details of our experimental setup are provided in the supplementary material.

\subsection{Quantitative Evaluation}
\paragraph{Robot Manipulation}
\begin{table*}[t]
    \centering
    \caption{Quantitative comparisons of robot manipulation tasks on SimplerEnv~\cite{li24simpler} and LIBERO~\cite{liu2023libero} benchmarks. \textbf{Bold} denotes the best result.}
    \resizebox{\textwidth}{!}{
        \label{tab:simpler_libero}
        \begin{tabular}{ll|cccccccc}
            \toprule
            \textbf{Dataset} & \textbf{Split} &
            \makecell{\textbf{Octo-Base}} &
            \makecell{\textbf{RT1-X}} &
            \makecell{\textbf{OpenVLA}} &
            \makecell{\textbf{DiT-Policy}} &
            \makecell{\textbf{TraceVLA}} &
            \makecell{\textbf{CoT-VLA}} &
            \makecell{\textbf{Magma}} &
            \textbf{\makecell{\ours{}\\(Ours)}} \\
            \midrule
            \multirow{4}{*}{\makecell[l]{Simpler-Google\\(Visual Matching)}}
              & Open/Close Drawer & 1.0 & 22.5 & 49.5 & 44.9 & 57.0 & -- & 56.0 & 50.0 \\
              & Move Near        & 3.0 & 55.0 & 47.1 & 58.9 & 53.7 & -- & 65.4 & 72.4 \\
              & Pick Coke Can    & 1.3 & 52.8 & 15.3 & 64.3 & 28.0 & -- & 83.7 & 92.0 \\
              \rowcolor{gray!20}
              & Overall          & 1.8 & 43.4 & 37.3 & 56.0 & 46.2 & -- & 68.4 & \textbf{71.5} \\
            \midrule
            \multirow{4}{*}{\makecell[l]{Simpler-Google\\(Variant Aggregation)}}
              & Open/Close Drawer & 22.0 & 56.0 & 22.5 & 35.5 & 31.0 & -- & 53.4 & 47.6 \\
              & Move Near        & 4.2  & 34.2 & 54.0 & 52.8 & 56.4 & -- & 65.7 & 63.8 \\
              & Pick Coke Can    & 17.0 & 54.0 & 52.8 & 56.4 & 60.0 & -- & 68.8 & 84.0 \\
              \rowcolor{gray!20}
              & Overall          & 14.4 & 48.1 & 43.1 & 48.2 & 49.1 & -- & {62.6} & \textbf{65.1} \\
            \midrule
            \multirow{5}{*}{\makecell[l]{Simpler-Bridge\\(Visual Matching)}}
              & Put Carrot on Plate    & 8.3 & 4.2 & 4.2 & 29.4 & -- & -- & 31.0 & 37.5 \\
              & Stack Blocks           & 0.0 & 0.0 & 0.0 & 0.0 & -- & -- & 12.7 & 8.7 \\
              & Put Spoon on Towel     & 12.5 & 0.0 & 8.3 & 34.5 & -- & -- & 37.5 & 58.3 \\
              & Put Eggplant in Basket & 43.1 & 0.0 & 45.8 & 65.5 & -- & -- & 60.5 & 70.8 \\
              \rowcolor{gray!20}
              & Overall               & 16.0 & 1.1 & 14.6 & 32.4 & -- & -- & {35.4} & \textbf{43.8} \\
            \midrule
            \multirow{5}{*}{LIBERO}
              & Spatial & 78.9 & -- & 84.7 & 82.6 & 84.6 & 87.5 & -- & 88.3 \\
              & Object  & 85.7 & -- & 88.4 & 84.7 & 85.2 & 91.6 & -- & 91.4 \\
              & Goal    & 84.6 & -- & 79.2 & 82.1 & 75.1 & 87.6 & -- & 87.1 \\
              & Long    & 51.1 & -- & 53.7 & 57.6 & 54.1 & 69.0 & -- & 70.9 \\
              \rowcolor{gray!20}
              & Overall & 75.1 & -- & 76.5 & 76.8 & 74.8 & {83.9} & -- & \textbf{84.4} \\
            \bottomrule
        \end{tabular}
    }
\end{table*}

To assess the effectiveness of \ours{} on robot manipulation task, we evaluate on SimplerEnv~\cite{li24simpler} and LIBERO~\cite{liu2023libero}. SimplerEnv~\cite{li24simpler} includes Google-VM (Visual Matching), Google-VA (Variant Aggregation), and Bridge-VM setups, introducing variations in color, material, lighting, and camera pose to evaluate model robustness. For the LIBERO~\cite{liu2023libero} benchmark, following prior works~\cite{kim24openvla, zhao2025cot}, we evaluate on the LIBERO-Spatial, LIBERO-Object, LIBERO-Goal, and LIBERO-Long subtasks to test model generalization across spatial layouts, object variations, goal diversity, and long-horizon planning.

As shown in Tab.~\ref{tab:simpler_libero}, on the SimplerEnv, incorporating our reasoning-guided visual plan latents allows \ours{} to outperform our baseline action model, DiT-Policy, by 15.5\%, 16.9\%, and 11.4\% on Google-VM, Google-VA, and Bridge-VM, respectively, achieving the highest overall scores of 71.5\%, 65.1\%, and 43.8\% against all methods. On the LIBERO benchmark, \ours{} achieves the best overall success rate of 84.4\%, outperforming DiT-Policy and recent state-of-the-art CoT-VLA~\cite{zhao2025cot}, verifying the effectiveness on diverse robotic manipulation settings.

\begin{table*}[t]
    \centering
    \caption{Quantitative comparisons of embodied reasoning tasks on EgoPlan-Bench2, RoboVQA, and OpenEQA benchmarks. Note that, Qwen2.5-VL* indicates fine-tuning the original Qwen2.5-VL using EgoPlan-IT~\cite{chen2023egoplan} and RoboVQA~\cite{sermanet2024robovqa} datasets. \textbf{Bold} denotes the best result.}
    \resizebox{\textwidth}{!}{
        \label{tab:embodied_reasoning}
        \begin{tabular}{ll|ccccccccc}
            \toprule
            \textbf{Dataset} & \textbf{Split / Metric} &
            \makecell{\textbf{\footnotesize GPT-4V}} & \makecell{\textbf{\footnotesize LLaVA-Video}} &
            \makecell{\textbf{\footnotesize InternVL2.5}} & \makecell{\textbf{\footnotesize InternVL3}} & \makecell{\textbf{\footnotesize NVILA}} &
            \makecell{\textbf{\footnotesize Qwen2.5-VL}} & \makecell{\textbf{\footnotesize Qwen2.5-VL*}} &
            \makecell{\textbf{\footnotesize Magma}} & \makecell{\textbf{\footnotesize \ours{}}\\\textbf{\footnotesize (Ours)}} \\
            \midrule
            \multirow{5}{*}{\makecell[l]{EgoPlan-\\Bench2}}
              & Daily life   & 36.7 & 38.0 & 36.2 & 38.5 & 35.8 & 31.4 & 47.9 & 32.1 & 50.1 \\
              & Work         & 27.7 & 29.9 & 28.7 & 32.9 & 28.7 & 26.7 & 46.3 & 25.7 & 49.8 \\
              & Recreation   & 33.9 & 39.0 & 34.4 & 36.1 & 37.2 & 29.5 & 44.3 & 34.4 & 44.8 \\
              & Hobbies      & 32.5 & 37.4 & 35.4 & 37.2 & 35.4 & 28.6 & 44.2 & 29.3 & 45.2 \\
              \rowcolor{gray!20}
              & Overall      & 32.6 & 35.5 & 33.5 & 36.2 & 33.7 & 29.1 & 45.7 & 29.8 & \textbf{48.2} \\
            \midrule
            \multirow{5}{*}{RoboVQA}
              & BLEU-1   & 32.2 & 35.4 & 40.5 & 44.3 & 42.7 & 47.8 & 65.3 & 38.6 & 69.1 \\
              & BLEU-2   & 26.5 & 32.1 & 33.3 & 36.5 & 39.7 & 41.2 & 57.3 & 31.5 & 61.8 \\
              & BLEU-3   & 24.7 & 30.0 & 29.6 & 31.6 & 37.6 & 36.2 & 52.2 & 28.1 & 56.0 \\
              & BLEU-4   & 23.9 & 29.0 & 27.5 & 28.9 & 36.1 & 33.7 & 48.0 & 26.7 & 52.4 \\
              \rowcolor{gray!20}
              & Overall  & 26.8 & 31.6 & 32.7 & 35.3 & 39.0 & 39.7 & 55.7 & 31.2 & \textbf{59.8} \\
            \midrule
            \multirow{8}{*}{OpenEQA}
              & Obj.\ State      & 63.2 & 69.1 & 70.2 & 68.9 & 66.1 & 63.2 & 62.4 & 59.9 & 70.0 \\
              & Obj.\ Recog.     & 43.4 & 42.6 & 47.2 & 49.1 & 49.5 & 46.2 & 45.2 & 43.8 & 47.2 \\
              & Func.\ Reason.   & 57.4 & 50.3 & 56.2 & 54.6 & 51.0 & 51.2 & 52.3 & 50.0 & 53.2 \\
              & Spatial          & 33.6 & 46.2 & 44.1 & 43.3 & 43.1 & 41.2 & 42.8 & 39.3 & 47.6 \\
              & Attri.\ Recog.   & 57.2 & 64.1 & 64.9 & 74.4 & 69.3 & 63.0 & 65.0 & 58.3 & 71.1 \\
              & World Know.      & 50.7 & 60.5 & 56.5 & 53.1 & 59.4 & 54.3 & 54.2 & 53.3 & 58.6 \\
              & Obj.\ Loc.       & 42.0 & 38.2 & 41.9 & 45.0 & 39.9 & 36.5 & 41.9 & 38.9 & 45.9 \\
              \rowcolor{gray!20}
              & Overall          & 49.6 & 53.0 & 54.4 & 55.5 & 54.0 & 50.8 & 52.0 & 49.1 & \textbf{56.2} \\
            \bottomrule
        \end{tabular}
    }
\end{table*}

\paragraph{Embodied Reasoning}

In Tab.~\ref{tab:embodied_reasoning}, we assess the reasoning capability of \ours{} in embodied scenarios on three benchmarks: EgoPlan-Bench2~\cite{qiu2024egoplan}, RoboVQA~\cite{sermanet2024robovqa}, and OpenEQA~\cite{majumdar2023openeqa}. EgoPlan-Bench2~\cite{qiu2024egoplan} measures multi-step planning in egocentric daily-life scenarios, while RoboVQA~\cite{sermanet2024robovqa} focuses on long-horizon reasoning in robotic manipulation. \ours{} outperforms the second-best method by 2.5\% and 4.1 BLEU score on these two benchmarks, demonstrating its strength in long-horizon and multi-step planning. Separately, OpenEQA~\cite{majumdar2023openeqa} measures zero-shot embodied understanding across diverse environments. The enhanced reasoning ability of \ours{} enables better generalization and scene comprehension, resulting in strong performance on this benchmark.

\begin{figure}[t!]
    \centering
    \includegraphics[width=1.0\linewidth]{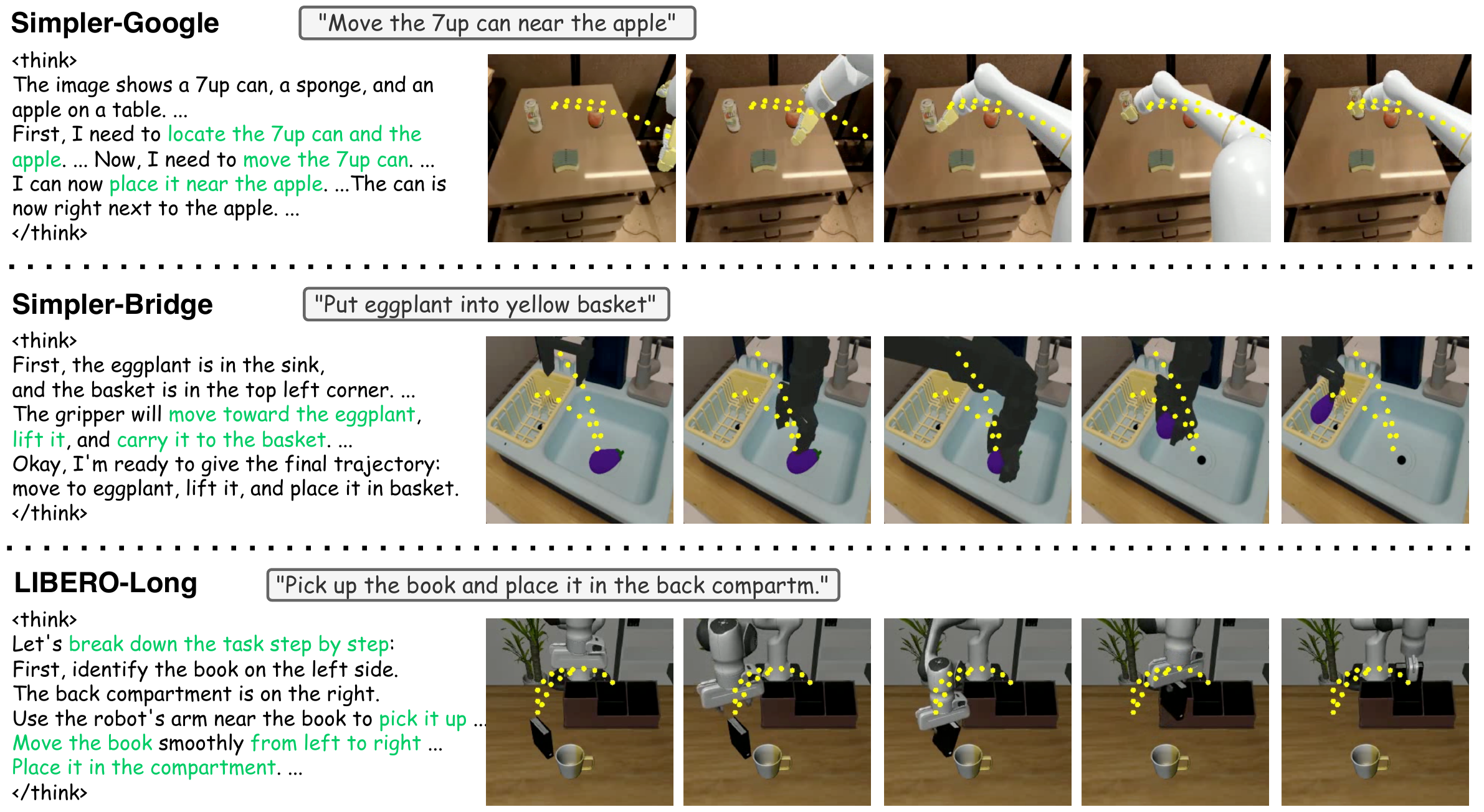}
    \caption{Qualitative results of intermediate reasoning steps and visualized trajectory for robot manipulation tasks on SimplerEnv and LIBERO benchmarks.}
    \label{fig:qualitative_manipulation}
\end{figure}

\begin{figure}[t!]
    \centering
    \includegraphics[width=1.0\linewidth]{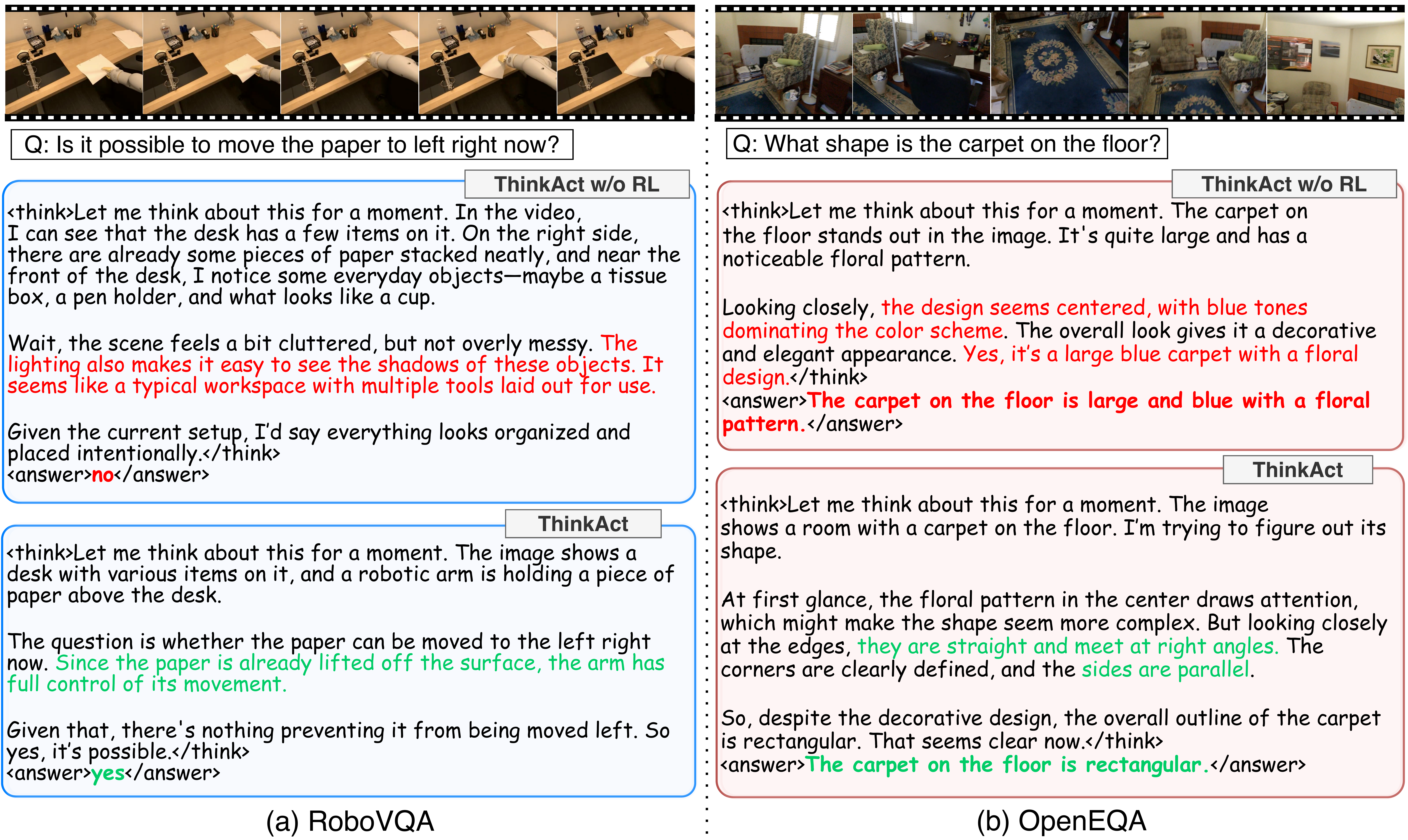}
    \caption{Qualitative comparison of reasoning process and the derived answer for our \ours{} with and without RL for embodied reasoning tasks on RoboVQA and OpenEQA benchmarks. \textcolor{red}{Red} denotes incorrect reasoning and answers, while \textcolor{green}{green} indicates correct ones.}
    \label{fig:qualitative_comparisons}
\end{figure}

\subsection{Qualitative Results}

In Fig.~\ref{fig:qualitative_manipulation}, we qualitatively showcase the reasoning process and execution scenes of two manipulation examples from the Simpler-Bridge~\cite{li24simpler} and LIBERO-Long~\cite{liu2023libero} tasks. In the LIBERO-Long task “Pick up the book and place it in the back compartment,” \ours{} decomposes the instruction into sub-tasks: (1) pick up the book, (2) move from left to right, and (3) place it in the compartment, demonstrating its \emph{long-horizon} planning capability. We also visualize the planned trajectory, confirming that the gripper closely follows the reasoning-guided plan during execution.

To better illustrate the impact of RL on the reasoning process, Fig.~\ref{fig:qualitative_comparisons} compares \ours{} before and after RL fine-tuning on embodied reasoning tasks. As we can observe in Fig.~\ref{fig:qualitative_comparisons}(a), using a RoboVQA~\cite{sermanet2024robovqa} example, the SFT cold-start model focuses only on the current state and fails to reason over future steps, while the RL-tuned model successfully infers the correct answer. Also, as demonstrated in Fig.~\ref{fig:qualitative_comparisons}(b), from OpenEQA~\cite{majumdar2023openeqa}, the cold-start model misinterprets the question, whereas the RL-tuned version demonstrates improved question and environment understanding. More qualitative comparisons and demo videos are provided in the supplementary material.

\subsection{Ablation Study}
In Tab.~\ref{tab:ablation_reward}, we ablate the proposed goal reward $r_\text{goal}$ and trajectory reward $r_\text{traj}$ to analyze their individual contributions to reasoning and planning. We start from the full version of \ours{}, which achieves the best performance across all benchmarks. Removing the trajectory reward leads to a noticeable drop, indicating that $r_\text{traj}$ is essential for learning coherent and structured planning behaviors. Without the goal reward, performance also declines, suggesting that $r_\text{goal}$ plays a key role in incentivizing long-horizon reasoning. When both $r_\text{traj}$ and $r_\text{goal}$ are removed, leaving only QA-style reward from QA datasets, the model shows only marginal improvements over the SFT baseline, confirming that action-aligned visual feedback is critical for effective multi-step planning in embodied settings. Finally, the SFT cold-start model without RL yields the lowest scores, verifying the effectiveness of our RL fine-tuning for eliciting the reasoning capability in MLLMs. More ablation studies (e.g., the number of interactions per reasoning step $N$) are provided in the supplementary material.

\subsection{Analysis of \ours{}}

In this section, we analyze the capabilities of \ours{} in enhancing robotic manipulation by embodied reasoning. We focus on two key aspects: (1) how reasoning facilitates effective few-shot adaptation to new tasks and environments, and (2) how it enables the robot to detect failures and perform self-correction during task execution. Through both quantitative experiments and qualitative examples, we demonstrate the unique advantages of leveraging a reasoning MLLM to tackle embodied action tasks. We further provide the analysis of MLLM backbones in the supplementary material.

\begin{figure}[t!]
    \centering
    \begin{minipage}{0.52\textwidth}
        \centering
        \captionof{table}{Quantitative ablation study for our proposed RL rewards in \ours{} on SimplerEnv, EgoPlan-Bench2, and RoboVQA benchmarks.}
        \resizebox{1.0\linewidth}{!}{
            \label{tab:ablation_reward}
            \begin{tabular}{lccccc}
                \toprule
                \textbf{Method} & \textbf{SimplerEnv} & \textbf{EgoPlan} & \textbf{RoboVQA}\\
                \midrule
                \textbf{\ours{} (Ours)} & \textbf{60.1} & \textbf{48.2} & \textbf{59.8} \\
                \midrule
                Ours w/o $r_{\text{traj}}$  & 59.2 & 47.9 & 58.5 \\
                Ours w/o $r_{\text{goal}}$ & 59.1 & 47.6 & 58.9 \\
                Ours w/o $r_{\text{traj}}, r_{\text{goal}}$ & 56.9 & 47.2 & 58.3 \\
                \midrule
                SFT cold-start & 56.4 & 46.4 & 57.9 \\
                \bottomrule
            \end{tabular}
        }
    \end{minipage}
    \hfill
    \begin{minipage}{0.45\textwidth}
        \centering
        \includegraphics[width=\linewidth]{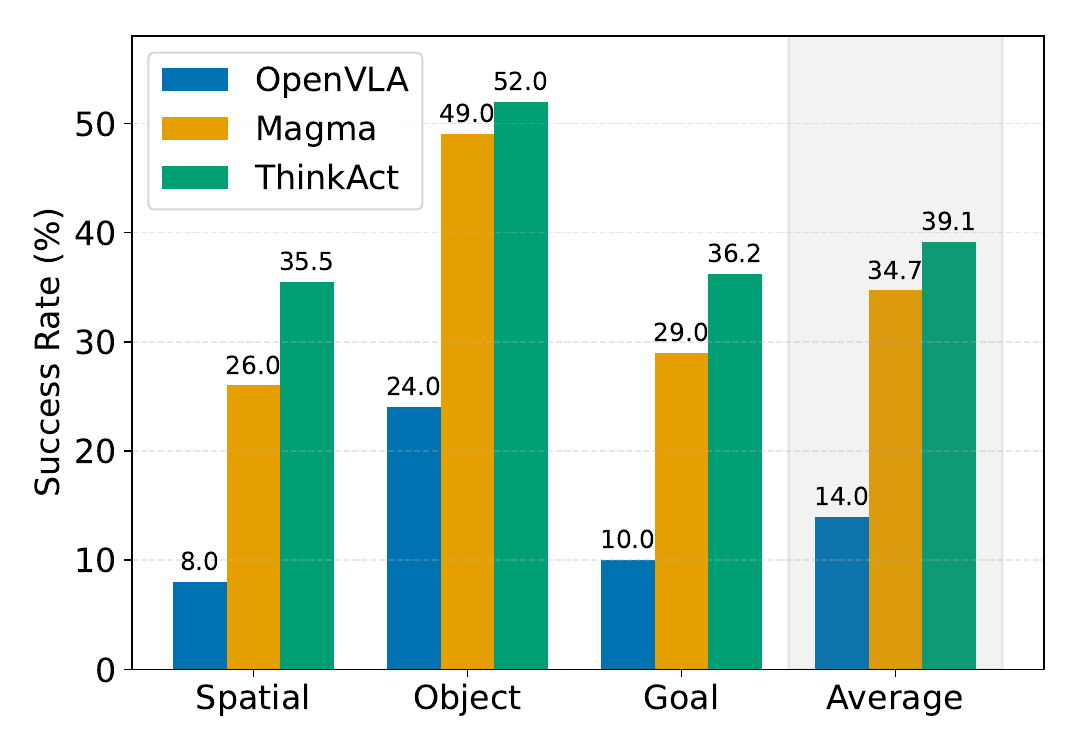}
        \vspace{-7mm}
        \caption{Few-shot adaptation results on LIBERO. We use 10 demonstrations per task for fine-tuning.}
        \label{fig:fewshot}
    \end{minipage}%
\end{figure}
\begin{figure}[t!]
    \centering
    \includegraphics[width=1.0\linewidth]{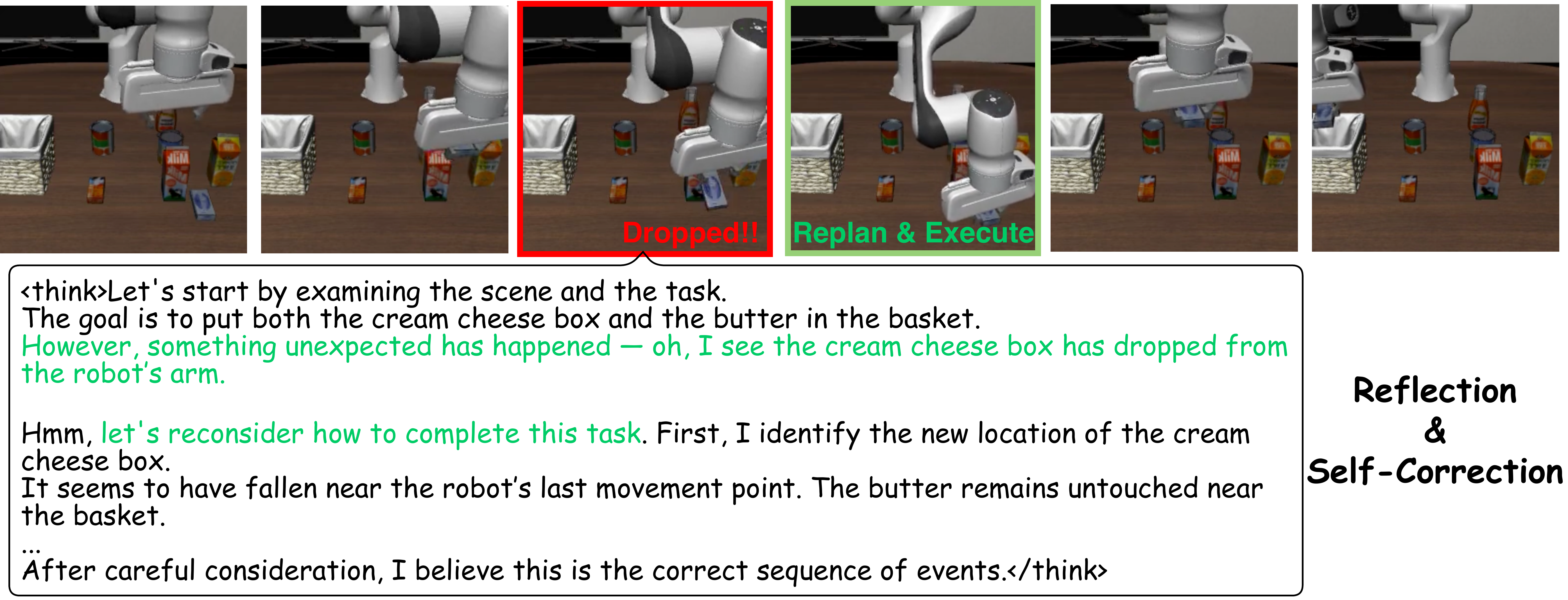}
    \caption{Demonstration of self-reflection and correction capability of \ours{}. The robot accidentally drops the target object midway. The reasoning MLLM identifies the failure and generates a revised plan that guides the gripper back to regrasp the object.}
    \label{fig:qualitative_reflection}
\end{figure}

\paragraph{Reasoning Enhance Few-Shot Adaptation}
As we can observe in Fig.~\ref{fig:qualitative_manipulation} and Fig.~\ref{fig:qualitative_comparisons}, \ours{} is capable of describing the environment and decomposing task instructions into meaningful sub-goals. To validate whether such reasoning improves the action model's adaptability, we conduct a few-shot adaptation experiment on the LIBERO benchmark~\cite{liu2023libero}. Specifically, we use LIBERO-Spatial and LIBERO-Object to evaluate adaptation to \textit{unseen environments}, and LIBERO-Goal to test adaptation to \textit{new skills}. We fine-tune the action model on just 10 demonstrations per task and evaluate performance over 100 trials. As shown in Fig.~\ref{fig:fewshot}, \ours{} consistently outperforms state-of-the-art methods, achieving the highest success rates across all tasks. Notably, it surpasses Magma~\cite{yang2025magma} by 7.3\% on LIBERO-Goal and by 9.5\% on LIBERO-Spatial, demonstrating the effectiveness of reasoning capability for few-shot generalization in both novel skills and environments.

\paragraph{Reasoning Elicit Self-Correction}
Failure detection and self-correction are critical for robust robot manipulation~\cite{liu2023reflect}. To evaluate whether \ours{} can reason about and recover from execution errors, we enable the reasoning MLLM to observe more contextual information during execution by extending its input from a single image $o_t$ to a short video segment $o_{t-N:t}$. This temporal context allows \ours{} to detect failures, reconsider the situation, and replan accordingly. For example, as shown in Fig.~\ref{fig:qualitative_reflection}, in a task where the robot is instructed to place a box into a basket, the gripper accidentally drops the box midway. The reasoning MLLM identifies the failure, says “Let's reconsider how to complete the task,” and generates a revised plan that guides the gripper back to the dropped location to regrasp the box. The robot then successfully completes the task, demonstrating \ours{}'s ability to reflect on errors and self-correct through structured reasoning.

\section{Conclusion}
\label{sec:conclusion}

We presented \textit{\ours{}}, a framework that reinforces visual latent planning for vision-language-action reasoning tasks. By combining action-aligned reinforcement learning with reasoning-enhanced action adaptation, \ours{} enables embodied agents to think before acting and execute robust actions in dynamic environments. Through extensive experiments across embodied reasoning and robot manipulation benchmarks, we demonstrated strong long-horizon planning, few-shot adaptation, and emergent behaviors such as failure detection and self-correction, providing a scalable path toward more deliberative and adaptable embodied AI systems.

\paragraph{Limitations}
Since ThinkAct builds on pretrained multimodal LLMs, it inevitably inherits their limitations, particularly hallucinations in visual or spatial reasoning. This can lead to generated plans that reference incorrect object attributes or spatial relationships, affecting downstream execution. While our latent planning and action grounding mitigate this to some extent, future work on grounding-aware training or hallucination suppression in MLLMs may further improve robustness and reliability in real-world deployment.

\paragraph{Broader Impacts}
Our work aims to enhance the reasoning capabilities of embodied agents, which could support real-world applications such as assistive robotics, home automation, and industrial systems. In particular, models like ThinkAct may help robots better interpret vague instructions and execute multi-step plans in dynamic environments. However, increased autonomy and reasoning ability in embodied systems also raise potential concerns. Misinterpretation of ambiguous commands, reliance on hallucinated visual reasoning, or overconfidence in CoT outputs could result in unintended behaviors, especially in safety-critical settings. Hence, future research on safeguards or alignment with human intent could further help mitigate these risks.

\clearpage
\appendix

\renewcommand{\thefigure}{A\arabic{figure}}
\renewcommand{\thetable}{A\arabic{table}}

\section{Additional Experimental Setup}

\subsection{Implementation Details}

\paragraph{Reinforced Fine-Tuning for Eliciting Visual Latent Planning} We set $\beta$ in GRPO to $1\text{e}{-2}$, with a maximum response length of 1024. To encourage diversity during rollout generation, we set the temperature to 1.0 and use top-$p$ sampling with $p = 0.99$. For computational efficiency, we use up to 16 video frames, each processed at a maximum resolution of $128 \times 28 \times 28$ pixels for video data, and $256 \times 28 \times 28$ pixels for image data. The length of trajectory, $K$, is set to 8, and for additional QA data, following~\cite{feng2025video}, we use accuracy as the reward for multiple-choice questions, and the average ROUGE-1/2/L scores for free-form answers.

\paragraph{Reasoning-Enhanced Action Adaptation} As mentioned in Sec.~4.1, the action model $\pi_\phi$ is a Transformer-based diffusion policy~\cite{chi2023diffusion}. We use a DDPM noise scheduler with 1000 timesteps for training, and inference using 20 DDIM steps. To accelerate training, for each observation $o_t$ and instruction $l$ pair, we let the MLLM $\mathcal{F}_\theta$ reason and generate the visual plan latent $c_t$ in an offline manner. With these cached latents, as described in Sec.~3.3, we train the action model $\pi_\theta$ via imitation learning while keeping the VLM frozen. We set the number of interactions per reasoning step $N$ to 15 for SimplerEnv~\cite{li24simpler} and 75 for the LIBERO benchmark~\cite{liu2023libero}, based on the average task length in each environment. We provide an ablation study on the choice of $N$ in Sec.~\ref{supp:ssec:ablation}. Following OpenVLA~\cite{kim24openvla}, we use a single $224 \times 224$ RGB image in third-person view as the observation input during training and inference.

\subsection{Training Data Preparation}

\subsubsection{Training Datasets}
\paragraph{2D Trajectory of Manipulation} Visual trajectories are sourced from two datasets: Open X-Embodiment (OXE)~\cite{o2024open} for robot manipulation, and Something-Something V2~\cite{goyal2017something} for human manipulation. Specifically, we select the fractal20220817\_data and bridge subsets from OXE for their high quality and visually clear trajectories. As described in Sec.~4.1, we extract gripper positions from each frame using an off-the-shelf detector\cite{niu2024llarva}. From each video, we randomly sample 3 starting frames and simplify the subsequent gripper trajectories into $K$ keypoints using the Ramer–Douglas–Peucker (RDP) algorithm (following HAMSTER~\cite{li2025hamster}). For Something-Something V2, we instead use a hand detector~\cite{Shan20}. In case two hands appear, we select the one with the largest movement. We apply stabilization~\cite{yang2025magma} to reduce the impact of camera motion.

\paragraph{RoboVQA~\cite{sermanet2024robovqa}} RoboVQA comprises a diverse set of real-world task episodes collected from both robotic and human embodiments. It contains approximately 5K long-horizon and 92K medium-horizon videos, each annotated with multiple question–answer pairs.

\paragraph{Reflect (RoboFail)~\cite{liu2023reflect}} The RoboFail dataset captures robot manipulation failures in both simulation and real-world scenarios. It includes 100 simulated failure cases in the AI2THOR environment and 30 real-world cases collected via UR5e teleoperation. We reformulate the original textual annotations into a multiple-choice question format, resulting in a total of 300 question–answer pairs.

\paragraph{EgoPlan-Bench~\cite{chen2023egoplan}} EgoPlan-Bench consists of egocentric videos annotated with task goals, progress histories, and current observations, designed to enhance MLLM planning capabilities in long-horizon daily tasks. It includes EgoPlan-IT, a 50K-instance subset generated automatically, and EgoPlan-Val, a 5K-instance, human-verified subset of high-quality samples.

\paragraph{Video-R1-CoT~\cite{feng2025video}} Video-R1-CoT comprises 165K question–answer samples with chain-of-thought (CoT) annotations generated by Qwen2.5-VL-72B~\cite{bai2025qwen2}. It is curated to support cold-start fine-tuning for video reasoning and spans domains including math, spatial logic, OCR, and chart understanding. All annotations are filtered for consistency and quality.

\paragraph{LLaVA-Video-178K~\cite{zhang2024video}} LLaVA-Video-178K includes 178K videos with detailed captions, 960K open-ended questions, and 196K multiple-choice questions. The annotations are generated via a GPT-4o-based pipeline, providing multi-level temporal descriptions and diverse question types, sourced from untrimmed videos across domains such as cooking, physical activities, and egocentric perspectives.

\subsubsection{Training Data Construction}
\paragraph{Supervised Fine-Tuning for Cold Start} For the SFT cold-start stage, we fine-tune the MLLM using 2D visual trajectories from OXE~\cite{o2024open}, QA tasks from RoboVQA~\cite{sermanet2024robovqa} and EgoPlan-IT~\cite{chen2023egoplan}, as well as chain-of-thought (CoT) data from Video-R1-CoT~\cite{feng2025video}. Specifically, the SFT dataset comprises 30K 2D visual trajectories, 50K RoboVQA samples, 50K EgoPlan-IT samples, and 165K Video-R1-CoT samples.

For the Video-R1-CoT data, which includes CoT annotations, we follow the original template~\cite{feng2025video}, prompting the model to output responses in the \texttt{<reason>}...\texttt{</reason>} \texttt{<answer>}...\texttt{</answer>} format. For the remaining datasets, which consist of standard QA pairs without intermediate reasoning, we append the instruction: ``Please directly provide your text answer within the \texttt{<answer>} \texttt{</answer>} tags, without any reasoning process,'' to encourage concise responses.

\paragraph{Reinforced Fine-Tuning for Eliciting Visual Latent Planning} For the reinforced fine-tuning stage, we use 2D visual trajectories from both OXE~\cite{o2024open} and Something-Something V2~\cite{goyal2017something}, along with QA datasets including RoboVQA~\cite{sermanet2024robovqa}, EgoPlan-IT/Val~\cite{chen2023egoplan}, RoboFail~\cite{liu2023reflect}, and LLaVA-Video-178K~\cite{li2024llava}. Specifically, the dataset consists of 12.5K 2D visual trajectories, 10K RoboVQA samples, 10K EgoPlan-IT/Val samples, 0.5K RoboFail samples, and 10K LLaVA-Video-178K samples.

We provide the detailed prompt templates for each data type in Tab.~\ref{tab:prompt_template}. This mixture of action-grounded and reasoning-intensive data enables the model to plan both physically executable and semantically coherent, while also improving generalization to diverse real-world tasks.

\begin{table}[ht]
    \centering
    \caption{Reasoning prompt template for reinforced fine-tuning.}\label{tab:prompt_template}
    \begin{tabular}{|p{2.5cm}|p{10cm}|}
        \hline
        \textbf{Data Type} & \textbf{Prompt Template} \\
        \hline
        2D Manipulation Trajectory & 
        Given an image of a robot manipulation scene and the task instruction "\{Instruction\}", please generate a sequence of 8 keypoints, representing the gripper's 2D trajectory on the image from its current position to the task-completion position. Please think about this planning process as if you were a human carefully reasoning through the manipulation task. Engage in an internal dialogue while considering the scene, the goal, possible subtasks, the motion path, and any obstacles. It's encouraged to include reflections on the environment, analysis of the goal state, decomposition into subtasks, and any adjustments to the planned trajectory as you think through the process. Provide your detailed reasoning between the <think> </think> tags, and then give your final prediction between the <answer> </answer> tags based on the reasoning. \newline
        Please provide the trajectory [(x1, y1), (x2, y2), ..., (x8, y8)] with coordinates normalized to [0,1] within <answer> </answer> tags.\\
        \hline
        QA Tasks & 
        \{Question\} Please think about this question as if you were a human pondering deeply. Engage in an internal dialogue using expressions such as 'let me think', 'wait', 'Hmm', 'oh, I see', 'let's break it down', etc, or other natural language thought expressions. It's encouraged to include self-reflection or verification in the reasoning process. Provide your detailed reasoning between the <think> </think> tags, and then give your final answer between the <answer> </answer> tags based on the reasoning. \newline
        (MCQ) Please provide only the single option letter (e.g., A, B, C, D, etc.) within the <answer> </answer> tags. \newline
        OR \newline
        (Free-form) Please provide your text answer within the <answer> </answer> tags.
         \\
        \hline
    \end{tabular}
\end{table}

\subsection{Evaluation Benchmarks}

\paragraph{SimplerEnv~\cite{li24simpler}} SimplerEnv is a simulation benchmark featuring two evaluation settings: visual matching and variant aggregation. It provides diverse manipulation scenarios across different lighting conditions, table textures, backgrounds, object distractors, and robot camera poses. Built on WidowX and Google Robot setups, SimplerEnv helps assess VLA robustness and the effectiveness of reasoning capability under varied visual conditions.

\paragraph{LIBERO~\cite{liu2023libero}} LIBERO is a simulation benchmark for evaluating generalization in robotic manipulation across four structured task suites, each targeting a distinct generalization challenge: spatial layout variation (LIBERO-Spatial), object diversity (LIBERO-Object), goal variation (LIBERO-Goal), and long-horizon planning with mixed variations (LIBERO-Long). Following prior work~\cite{zhao2025cot}, we evaluate each task suite over 500 trials using 3 random seeds.

\paragraph{EgoPlan-Bench2~\cite{qiu2024egoplan}}
EgoPlan-Bench2 evaluates the egocentric planning capabilities of MLLMs in complex, real-world scenarios. It emphasizes long-horizon reasoning based on task goals, progress, and current observations, spanning 24 scenarios across 4 daily-life domains. Compared to EgoPlan-Bench~\cite{chen2023egoplan}, it features more diverse scenes and serves as a non-overlapping evaluation set. The benchmark includes 1,321 high-quality multiple-choice QA pairs evaluated using accuracy.

\paragraph{RoboVQA~\cite{sermanet2024robovqa}}
RoboVQA focuses on visual question answering in robotic manipulation, emphasizing long-horizon reasoning, contextual understanding, and affordance-based decision-making. It includes real-world videos from both robot and human embodiments, covering planning, future prediction, affordance reasoning, and outcome classification. We use its validation set, which consists of 1,893 video–text pairs in a free-form QA format evaluated using the BLEU score.

\paragraph{OpenEQA~\cite{majumdar2023openeqa}}
OpenEQA is a benchmark for embodied question answering (EQA), aiming to evaluate an agent's ability to understand and reason about real-world environments through natural language. It poses questions that require spatial, functional, and commonsense understanding across diverse scenes. The dataset includes over 1,600 high-quality human-authored questions from more than 180 real-world environments, in a free-form QA format evaluated using an LLM-based scoring metric aligned with human judgment.

\section{Additional Experiment Results}

\begin{figure}[ht]
    \centering
    \includegraphics[width=1.0\linewidth]{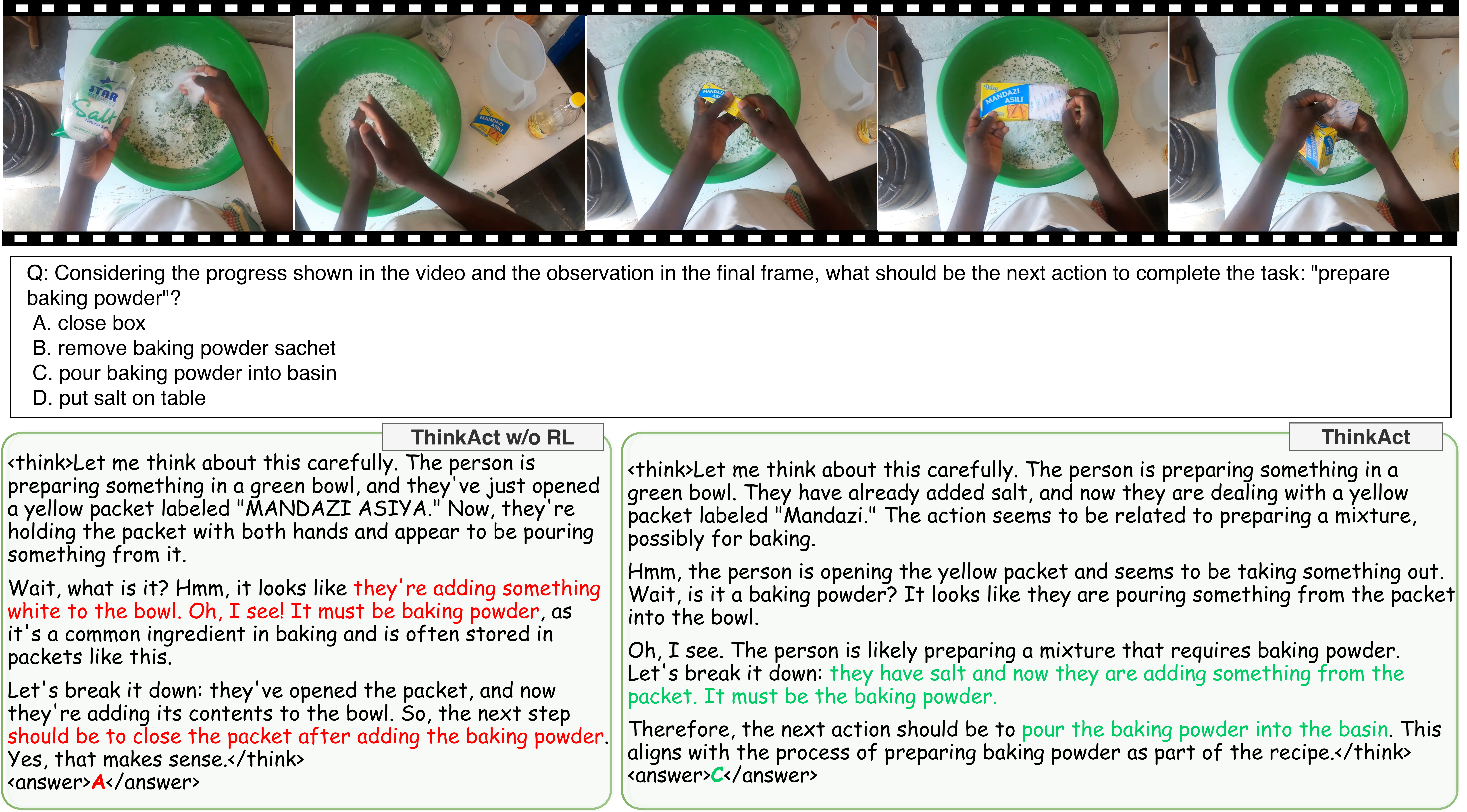}
    \caption{Qualitative comparison of reasoning process and the derived answer for our \ours{} with and without RL for embodied reasoning tasks on EgoPlan-Bench2 benchmark. \textcolor{red}{Red} denotes the incorrect reasoning and answer, while \textcolor{green}{green} indicates the correct one.}
    \label{fig:qualitative_comparisons_egoplan}
\end{figure}

\subsection{Qualitative Comparisons of Robot Execution Results}
To complement the quantitative results, we provide qualitative comparisons of robot execution results between DiT-Policy~\cite{chi2023diffusion}, OpenVLA~\cite{kim24openvla}, and \ours{} in the supplementary video file \texttt{ThinkAct.mp4}.

\subsection{Additional Qualitative Results}
Fig.~\ref{fig:qualitative_comparisons_egoplan} presents a comparison of ThinkAct before and after RL fine-tuning on an EgoPlan-Bench2~\cite{qiu2024egoplan} example.  Similar to Fig.~4 in the main paper, RL enhances embodied reasoning, enabling the model to predict the correct next action.

\subsection{More Self-Correction Samples}
\begin{figure}[ht]
    \centering
    \includegraphics[width=1.0\linewidth]{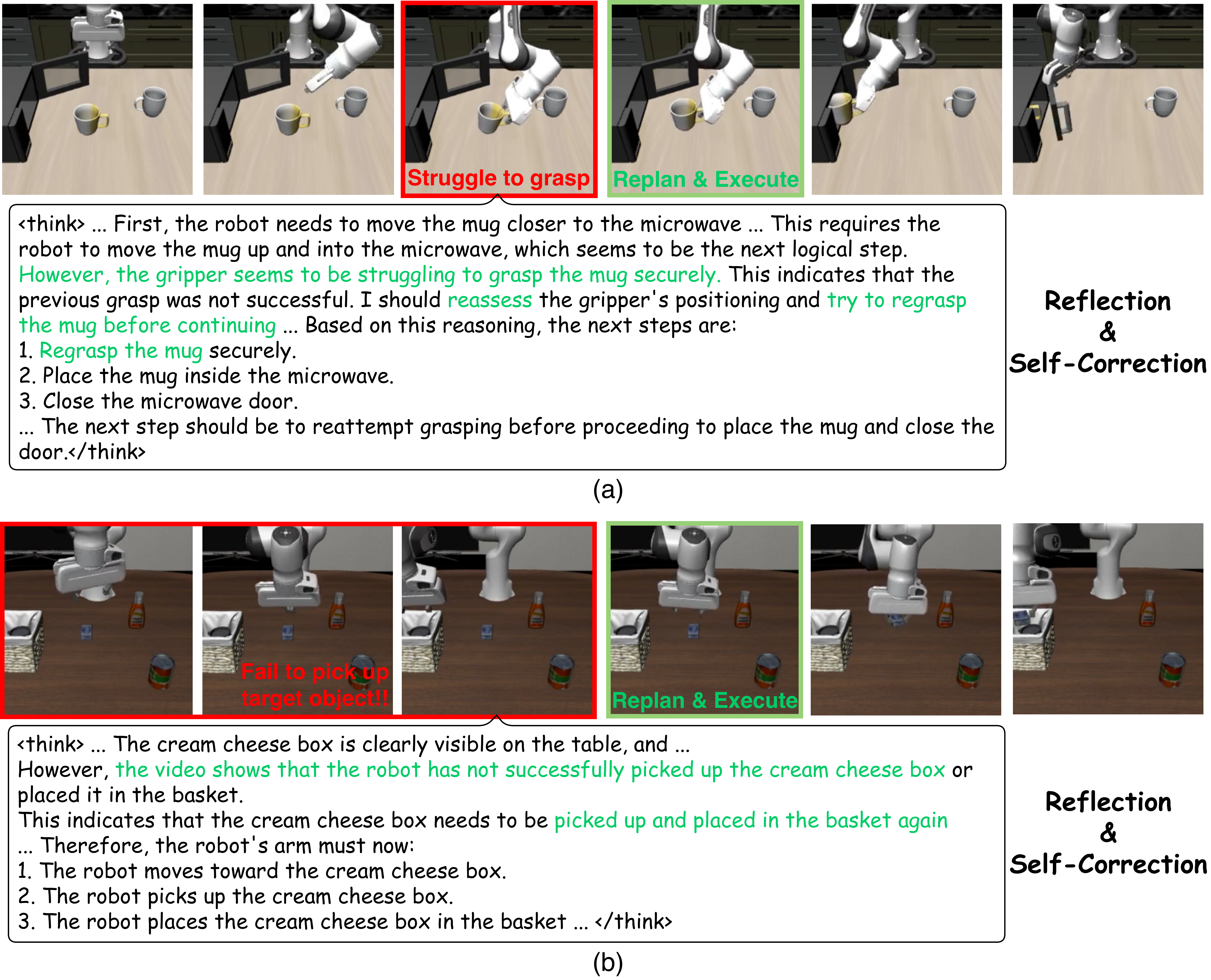}
    \caption{More Demonstrations of self-reflection and correction capability of ThinkAct.}
    \label{fig:reflect_more}
\end{figure}

To further demonstrate the capacity of \ours{} for reflection and self-correction, we present two additional examples. In Fig.~\ref{fig:reflect_more}(a), the robot fails to grasp a mug. The reasoning MLLM identifies the issue, noting that the gripper is struggling, and suggests adjusting its position to reattempt the grasp. In Fig.~\ref{fig:reflect_more}(b), the robot attempts to move an object to a basket, but fails to pick it up in the first place. The MLLM detects the failure and replans the pickup, leading to successful completion. These cases highlight \ours{}’s ability to detect and recover from execution errors through reasoning.

\subsection{Results of Smaller Model Size} To demonstrate the generalizability of our approach, we apply \ours{} to a smaller model, Qwen2.5-VL-3B, and compare its performance with other models of similar size. As shown in Tab.~\ref{tab:supp_embodied_reasoning}, \ours{} consistently outperforms other models on EgoPlan-Bench2~\cite{qiu2024egoplan}, RoboVQA~\cite{sermanet2024robovqa}, and OpenEQA~\cite{majumdar2023openeqa}, demonstrating its effectiveness on smaller MLLM backbone.

\begin{table*}[ht]
    \centering
    \caption{Quantitative comparisons with smaller models on embodied reasoning tasks.}
    \resizebox{\textwidth}{!}{
    \label{tab:supp_embodied_reasoning}
    \begin{tabular}{ll|cccccc}
        \toprule
        \textbf{Dataset} & \textbf{Split / Metric} &
        \makecell{\textbf{\footnotesize InternVL2.5-2B}} &
        \makecell{\textbf{\footnotesize InternVL3-2B}} &
        \makecell{\textbf{\footnotesize NVILA-2B}} &
        \makecell{\textbf{\footnotesize Qwen2.5-VL-3B}} &
        \makecell{\textbf{\footnotesize Qwen2.5-VL-3B}*} &
        \makecell{\textbf{\footnotesize ThinkAct-3B}\\\textbf{\footnotesize (Ours)}} \\
        \midrule
        \multirow{5}{*}{\makecell[l]{EgoPlan-\\Bench2}}
          & Daily life   & 30.9 & 36.9 & 34.6 & 29.0 & 44.9 & 46.6 \\
          & Work         & 27.8 & 29.9 & 26.7 & 27.0 & 43.0 & 41.4 \\
          & Recreation   & 28.6 & 35.6 & 33.3 & 30.2 & 42.2 & 45.9 \\
          & Hobbies      & 33.1 & 31.5 & 31.6 & 28.9 & 40.9 & 42.5 \\
          \rowcolor{gray!20}
          & Overall      & 30.1 & 33.4 & 31.4 & 28.5 & 43.0 & \textbf{44.0} \\
        \midrule
        \multirow{5}{*}{RoboVQA}
          & BLEU-1   & 36.6 & 34.4 & 38.7 & 42.5 & 60.7 & 62.4 \\
          & BLEU-2   & 33.7 & 33.9 & 34.3 & 36.3 & 56.8 & 57.3 \\
          & BLEU-3   & 31.0 & 33.5 & 31.1 & 28.7 & 51.3 & 52.0 \\
          & BLEU-4   & 29.4 & 33.3 & 29.2 & 31.8 & 45.7 & 49.6 \\
          \rowcolor{gray!20}
          & Average  & 32.7 & 33.8 & 33.3 & 34.8 & 53.6 & \textbf{55.3} \\
        \midrule
        \multirow{8}{*}{OpenEQA}
          & Obj.\ State      & 60.5 & 61.2 & 59.7 & 59.8 & 56.3 & 60.6 \\
          & Obj.\ Recog.     & 43.7 & 42.8 & 39.6 & 37.8 & 41.7 & 45.3 \\
          & Func.\ Reason.   & 49.0 & 53.5 & 47.2 & 48.0 & 45.3 & 51.4 \\
          & Spatial          & 36.9 & 38.9 & 36.5 & 32.8 & 36.2 & 39.4 \\
          & Attri.\ Recog.   & 63.5 & 62.6 & 61.5 & 57.6 & 56.6 & 61.7 \\
          & World Know.      & 42.3 & 45.2 & 51.3 & 38.9 & 40.9 & 46.4 \\
          & Obj.\ Loc.       & 33.6 & 37.2 & 33.1 & 29.0 & 35.3 & 37.6 \\
          \rowcolor{gray!20}
          & Overall          & 47.1 & 48.8 & 47.0 & 43.4 & 44.6 & \textbf{48.9} \\
        \bottomrule
    \end{tabular}}
\end{table*}

\subsection{Results of 5-Shot Adaptation}
As shown in Fig.~\ref{fig:5shot}, we conduct an additional 5-shot adaptation experiment on LIBERO~\cite{liu2023libero}. Specifically, we fine-tune the action model using only 5 demonstrations per task and evaluate its performance over 100 trials, following the protocol of Magma~\cite{yang2025magma}. Consistent with the 10-shot results in Fig.~5 of the main paper, \ours{} consistently outperforms comparative methods across all three tasks.

\subsection{Ablation Study}\label{supp:ssec:ablation}
\begin{figure}[ht]
    \centering
    \begin{minipage}{0.43\textwidth}
        \centering
        \captionof{table}{Quantitative ablation study for our proposed RL rewards in \ours{} on LIBERO and OpenEQA benchmarks.}
        \resizebox{1.0\linewidth}{!}{
            \label{tab:supp_ablation_reward}
            \begin{tabular}{lcccc}
                \toprule
                \textbf{Method} & \textbf{LIBERO} & \textbf{OpenEQA}\\
                \midrule
                \textbf{\ours{} (Ours)} & \textbf{84.4} & \textbf{56.2}\\
                \midrule
                Ours w/o $r_{\text{traj}}$  & 82.1 & 55.9 \\
                Ours w/o $r_{\text{goal}}$ & 81.7 & 55.6 \\
                Ours w/o $r_{\text{traj}}, r_{\text{goal}}$ & 81.6 & 55.7 \\
                \midrule
                SFT cold-start & 79.1 & 53.3 \\
                \bottomrule
            \end{tabular}
        }
    \end{minipage}
    \hfill
    \begin{minipage}{0.5\textwidth}
        \centering
        \includegraphics[width=\linewidth]{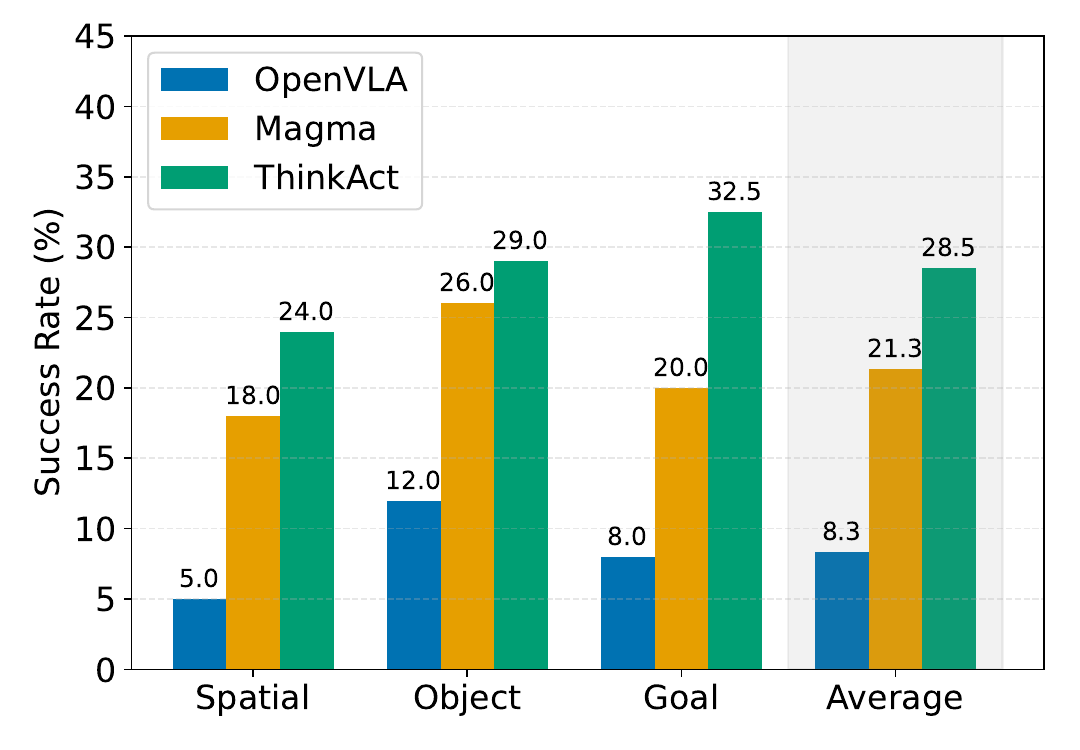}
        \vspace{-7mm}
        \caption{5-shot adaptation results on LIBERO.}
        \label{fig:5shot}
    \end{minipage}%
\end{figure}

\paragraph{Additional Quantitative Ablation on LIBERO and OpenEQA Benchmarks}
Tab.~\ref{tab:supp_ablation_reward} extends the main paper's ablation by evaluating on LIBERO~\cite{liu2023libero} and OpenEQA~\cite{majumdar2023openeqa}. Results confirm that both $r_{\text{goal}}$ and $r_{\text{traj}}$ are crucial for effective planning, with performance dropping when either is removed and nearing the SFT baseline when both are excluded. This further supports the importance of action-aligned visual rewards.

\paragraph{Ablation Study on the Number of Actions per Reason}
We ablate the frequency of reasoning updates by varying the number of actions per reasoning step $N$ on LIBERO. Setting $N$ to 25, 50, 75, and 100 results in average success rates of 84.0\%, 84.6\%, 84.4\%, and 83.7\%, respectively. These results suggest that overly sparse reasoning (e.g., $N{=}100$) might cause the model to be unable to detect the failure and perform self-correction in time, leading to degraded performance. On the other hand, too frequent updates (e.g., $N{=}25$) would induce additional inference cost without yielding substantial performance gains. As a result, we set the number of actions per reasoning $N$ as 75 on LIBERO.

\subsection{Inference Speed}
We compare the inference speed of \ours{} with the end-to-end OpenVLA~\cite{kim24openvla} on LIBERO~\cite{liu2023libero} tasks using an A100 GPU. On average, \ours{} takes 17\% longer execution time than OpenVLA, primarily due to the autoregressive reasoning process. 
We note that while the inference time slightly increases, our embodied reasoning, as a test-time scaling paradigm, significantly boosts downstream task performance.
That is, \ours{} outperforms OpenVLA on all four LIBERO task categories, achieving success rate improvements of 2.8\% on spatial, 3.2\% on object, 8.4\% on goal, and 15.3\% on long-horizon tasks. These results show that the reasoning overhead is justified by significant performance gains, highlighting the effectiveness of embodied reasoning for robot manipulation.

\clearpage
\setcitestyle{numbers}
\bibliographystyle{plainnat}
\bibliography{main}

\end{document}